\documentclass [conference] {IEEEtran}
\IEEEoverridecommandlockouts
\usepackage[colorlinks,urlcolor=blue,linkcolor=blue,citecolor=blue]{hyperref}
\usepackage{color,array}
\usepackage{graphicx}
\usepackage{algorithm}
\usepackage{algorithmic}
\makeatother
\usepackage[utf8]{inputenc}
\usepackage[table]{xcolor}
\usepackage{tabularx,booktabs}
\newcolumntype{C}{>{\centering\arraybackslash}X} 
\usepackage{cite}
\usepackage{amsmath}
\usepackage{amsmath,amssymb,amsfonts}
\usepackage{algorithmic}
\usepackage{graphicx}
\usepackage{textcomp}
\usepackage{xcolor}
\usepackage{hyperref}
\usepackage{caption}
\usepackage{pifont}
\usepackage{xcolor}
\usepackage{rotating}
\usepackage{rotfloat}
\usepackage{algorithm}
\usepackage{algorithmic}
\usepackage{gensymb}
\usepackage{textcomp}
\usepackage{placeins}
\usepackage{balance}
\usepackage{subfigure}
\usepackage{booktabs}
\usepackage{tabularx}
\usepackage{array}
\usepackage{pbox}
\usepackage{longtable}
\usepackage{booktabs}
\def\BibTeX{{\rm B\kern-.05em{\sc i\kern-.025em b}\kern-.08em
    T\kern-.1667em\lower.7ex\hbox{E}\kern-.125emX}}

\makeatletter
\def\endthebibliography{%
  \def\@noitemerr{\@latex@warning{Empty `thebibliography' environment}}%
  \endlist
}
\makeatother

\begin{document}

\title{Gender Bias in Transformer Models: A comprehensive survey} 
\author{Praneeth Nemani, Yericherla Deepak Joel, Palla Vijay, Farhana Ferdouzi Liza}


\maketitle
\begin{abstract}
Abstract:
Gender bias in artificial intelligence (AI) has emerged as a pressing concern with profound implications for individuals' lives. This paper presents a comprehensive survey that explores gender bias in Transformer models from a linguistic perspective. While the existence of gender bias in language models has been acknowledged in previous studies, there remains a lack of consensus on how to effectively measure and evaluate this bias. Our survey critically examines the existing literature on gender bias in Transformers, shedding light on the diverse methodologies and metrics employed to assess bias. Several limitations in current approaches to measuring gender bias in Transformers are identified, encompassing the utilization of incomplete or flawed metrics, inadequate dataset sizes, and a dearth of standardization in evaluation methods. Furthermore, our survey delves into the potential ramifications of gender bias in Transformers for downstream applications, including dialogue systems and machine translation. We underscore the importance of fostering equity and fairness in these systems by emphasizing the need for heightened awareness and accountability in developing and deploying language technologies. This paper serves as a comprehensive overview of gender bias in Transformer models, providing novel insights and offering valuable directions for future research in this critical domain.
\end{abstract}

\begin{IEEEkeywords}
Gender bias, Transformer models, Bias mitigation, Binary gender assumption, Self-attention
\end{IEEEkeywords}

\section{Introduction}
\label{sec:introduction}
Artificial intelligence (AI) is often perceived as a neutral entity. However, as AI is created by humans, it reflects our prejudices, including gender bias \cite{nadeem2020gender, nadeem2022gender}. There is a growing concern in both the scientific community and the general public about the demographic biases in some AI applications \cite{schwartz2022towards}. Gender bias can perpetuate harmful stereotypes and biases, leading to unfair treatment and discrimination. This bias can also limit opportunities for marginalized groups, particularly in areas such as employment, education, and healthcare. Furthermore, it can negatively impact the accuracy and fairness of natural language processing (NLP) applications, affecting the user experience and reliability of these systems. As NLP and machine learning (ML) tools gain prominence, it is becoming increasingly important to understand how they contribute to the formation of societal prejudices and preconceptions. NLP models are effective at modeling many different applications, but they can reinforce gender prejudice that is present in text corpora. Despite the fact that gender bias in NLP has been studied since the 90s \cite{baldwin1995university, ozieblowska1994generic}, there is still much to learn about certain aspects. Therefore, it is crucial to examine how current research in NLP is establishing an entirely new field \cite{wang2019implicit}. To determine whether these studies are heading in the right direction to solve the problem, it is essential to evaluate whether they present scalable evaluation techniques and whether their objectives are well-stated. Bias refers to an unfair opinion or preference held in favor of or against a specific person or group \cite{simundic2013bias}. In machine learning, bias can be caused by faulty assumptions in the algorithm or systemic prediction errors caused by the characteristics of the training data \cite{turney1995bias, mehrabi2021survey}. Gender bias, which favors or stereotypes one gender over another, particularly males over females, has been studied extensively in relation to NLP applications. Different researchers have proposed various definitions and inferences of gender bias and its impact on NLP. 

\subsection{Gender Bias in Word Embeddings}

Word embeddings are crucial in transformer-based NLP models. They are a fundamental component that helps transformers understand and represent the meaning of words in a way that machines can process. Word embeddings are dense vector representations of words that capture semantic and syntactic relationships between them. These vectors encode contextual information about words based on their co-occurrence patterns in large text corpora. In transformer models, such as the popular architecture known as BERT (Bidirectional Encoder Representations from Transformers), word embeddings are used as input representations for the model. These embeddings provide the initial understanding of individual words within the context of a given sentence or document. The transformer model then processes these word embeddings through self-attention mechanisms, enabling it to capture contextual relationships between words. The attention mechanism allows the model to weigh the importance of each word in the context of the entire sentence, which is crucial for understanding the meaning of a word based on its surrounding words. The ability of word embeddings to capture semantic relationships allows transformer models to perform a wide range of NLP tasks effectively. These tasks include text classification, named entity recognition, sentiment analysis, machine translation, question answering, and many others. By leveraging word embeddings, transformers can learn and generalize patterns from vast amounts of text data, improving performance in various NLP tasks. The Artificial Intelligence and Emerging Technology Initiative of The Brookings Institution has explored the issue of bias and NLP research in their \textbf{"AI and Bias"} series, with a particular focus on gender bias. The author of one 2021 essay draws on their previous research at Princeton University's Center for Information Technology Policy, where they found that machine learning algorithms processing word embeddings can pick up biases similar to those of humans from the word associations in their training data \cite{brunet2019understanding, papakyriakopoulos2020bias, zhao2019gender}. One example of gender bias in word embeddings is the association of certain professions with gender. For instance, the word "nurse" may be more closely associated with the female gender than the male gender in a word embedding model. Similarly, "engineer" may be more closely associated with the male gender than the female gender. This bias can manifest in natural language processing applications such as automated resume screening. 

\begin{figure}[htbp]
    \centering
    \includegraphics[width = \linewidth]{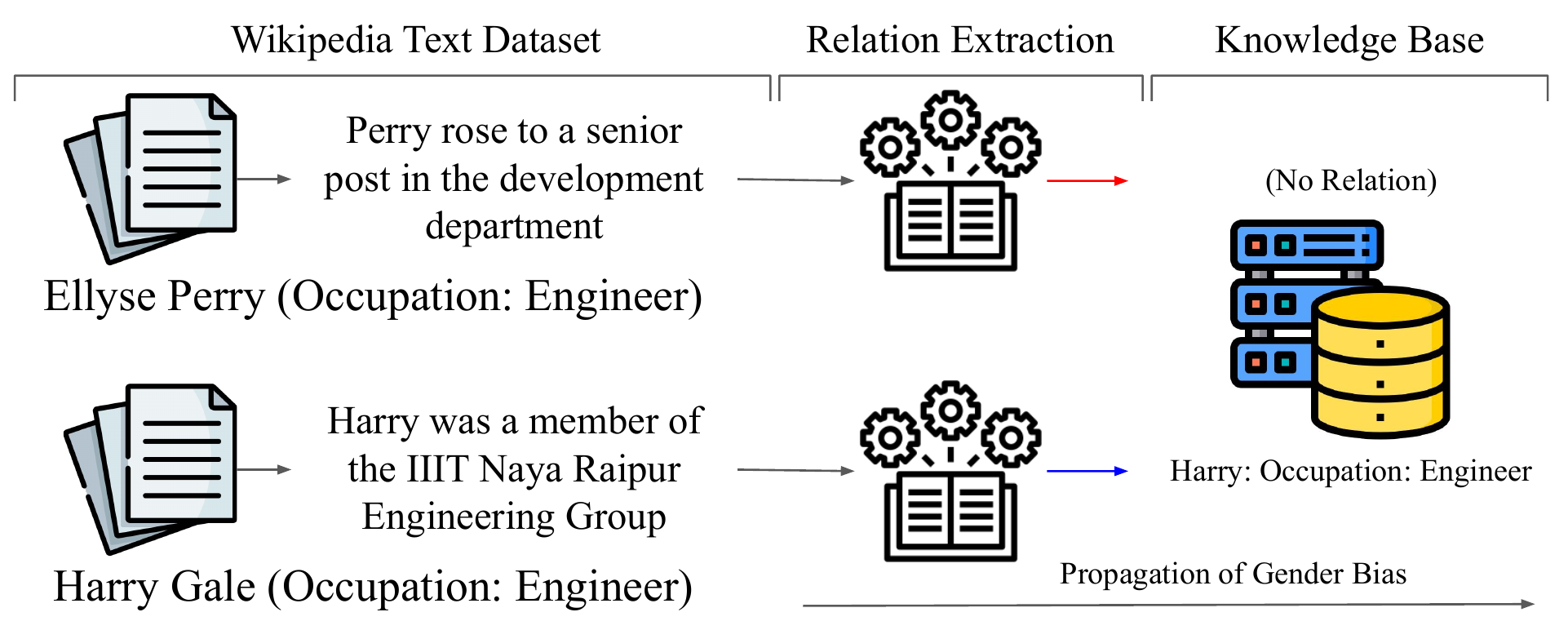}
    \caption{Gender Bias in Word Embeddings}
    \label{fig:WordBias}
\end{figure}

Another example of gender bias is the case of \textbf{Amazon \cite{WinNT}}. In 2018, it was reported that Amazon had developed an AI-powered hiring tool that was trained on resumes submitted to the company over a 10-year period. The tool was designed to screen resumes and rank candidates based on their qualifications, with the goal of identifying top candidates more efficiently. However, the tool was found to have a significant gender bias. This was because the tool was trained on resumes submitted to Amazon over the past decade, which were predominantly from men due to the tech industry's gender imbalance. As a result, the tool learned to associate certain words and phrases with male candidates and would downgrade resumes that contained language associated with women, such as references to women's colleges or women's sports teams. Amazon ultimately abandoned the AI-powered hiring tool, recognizing that it was not effective and potentially harmful due to its gender bias. The case highlighted the importance of ensuring that automated tools used in hiring and other applications are free from bias and trained on diverse datasets to ensure fairness and accuracy. To describe this representation professionally, one could say that Fig. \ref{fig:WordBias} presents a conceptual illustration of the way in which gender bias can manifest in word embeddings.

\subsection{Gender Bias in Machine Translation}

Gender bias in machine translation refers to the phenomenon where machine translation systems produce translations that reflect or reinforce gender stereotypes or where translations are inaccurate or inappropriate because of gender-related differences in the source and target languages \cite{stanovsky2019evaluating, prates2020assessing}. One common example of gender bias in machine translation is the use of gendered pronouns. Many languages, such as French and Spanish, use gendered pronouns to refer to people, and machine translation systems may struggle to accurately translate sentences that contain these pronouns. For example, the French sentence \textit{Le médecin a vu la patiente} can be translated into English as \textit{The doctor saw the patient} or \textit{The doctor saw the female patient}, but machine translation systems may default to using the masculine pronoun "he" instead of "she" when translating the sentence into English.

\begin{figure}[ht]
    \centering
    \includegraphics[width = \linewidth]{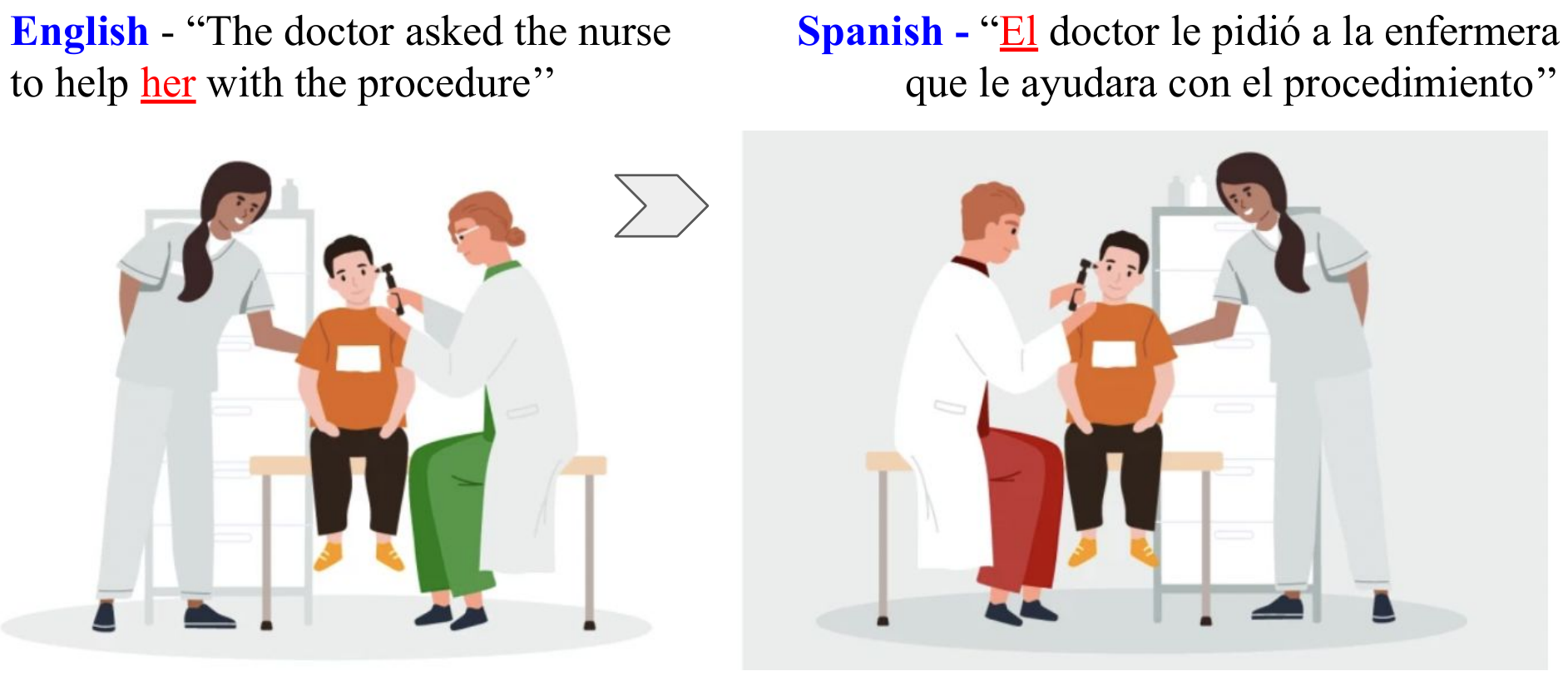}
    \caption{Evidence of Gender Bias in MT even due to the presence of unambiguous gender context}
    \label{fig:MTBias}
\end{figure}

Another example of gender bias in machine translation is the use of gender stereotypes in translations. For instance, a machine translation system may translate a sentence like \textit{She is a doctor} into a language where the word for doctor is masculine by default, resulting in a translation that reinforces the stereotype that doctors are male. Similarly, a system may translate a sentence like "He is a nurse" into a language where the word for nurse is feminine by default, which could be seen as reinforcing the stereotype that nurses are female. MIT press has suggested that the weight of prejudices and stereotypes, as well as the presence of equal gender weightage, can be used to assess gender bias in MT \cite{10.1162/tacl_a_00401}. In cases where a language with limited or no gender (such as English) is being interpreted into a language with significant grammatical gender, an ideal MT model should accurately translate and express the genders of words lacking gender in the input language (such as Spanish). Fig. \ref{fig:MTBias} visually illustrates such a bias in the translation from English to Spanish.

\subsection{Gender Bias in Caption Generation}

\begin{figure}[ht]
    \centering
    \includegraphics[width = \linewidth]{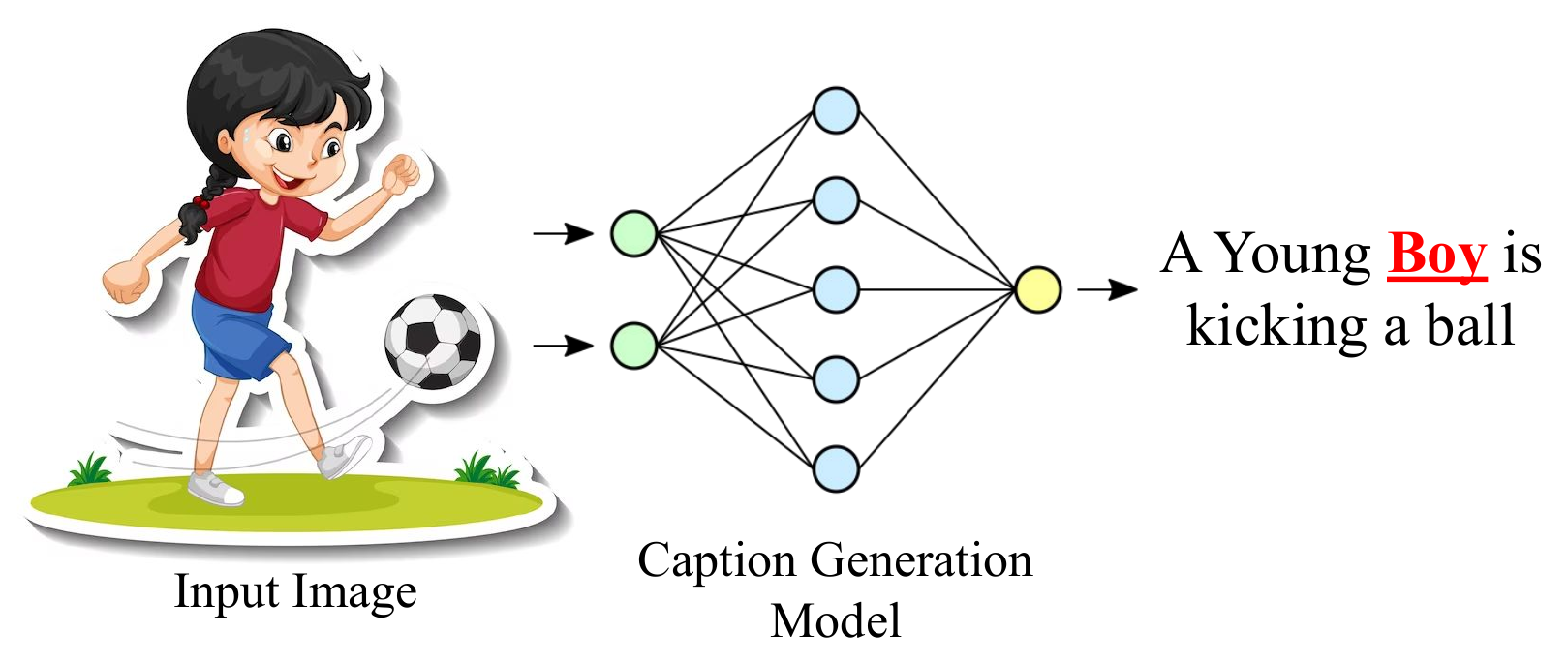}
    \caption{Biased scenario of caption generation}
    \label{fig:CapBias}
\end{figure}

Caption generation generates a textual description that accurately describes the content of an image or a video. Gender bias can also manifest in caption generation, where the generated captions can be biased towards a particular gender. The issue of gender bias is not limited to machine translation and word embeddings; it can also be found in caption generation tasks. Tang et al. \cite{Tang} conducted a study and found that captioning datasets, such as the COCO dataset, may lead to unintentional gender-biased models due to intrinsic memorization. The COCO training dataset exhibits a significant gender bias, which makes the 3:1 male-to-female ratio in the dataset even more unbalanced. An ideal model should not identify a person as a woman based on the background of a house, for instance. Instead, an unbiased model should predict gender terms based on visual characteristics associated with the depicted individual. The problem of gender bias in caption generation is depicted in Fig. \ref{fig:CapBias}, where the model predicts the output based on biased data. Other studies, such as \cite{hendricks2018women, hirota2022quantifying}, also highlight the issue of bias in caption generation.

\subsection{Gender Bias in Sentiment Analysis}
Sentiment analysis, which involves the automated identification of emotions or attitudes in large datasets, is a popular application of NLP and ML techniques. However, as pointed out by \textbf{Thelwall et al. \cite{thelwall2018gender}}, this task can be susceptible to bias due to under-representation. In this study, the researchers investigated whether biases exist in the accuracy of lexical sentiment analysis when applied to reviews written by individuals of different genders. Specifically, the analysis focused on TripAdvisor reviews of hotels and restaurants in the UK, authored by UK residents. The aim was to compare the effectiveness of lexical sentiment analysis in detecting sentiments expressed by males and females. The study's results revealed that detecting sentiment in reviews written by males was more challenging than those written by females. This difficulty stemmed from the fact that male sentiment tended to be less explicit or overtly expressed. Furthermore, the researchers found no evidence to support the notion that gender-specific lexical sentiment analysis could effectively address this issue. Similarly, studies by \textbf{Asyrofi et al. \cite{9653830}} and \textbf{Kiritchenko et al. \cite{kiritchenko2018examining}} have also highlighted the issue of gender bias in sentiment analysis. In their work, \textbf{Asyrofi et al. \cite{9653830}} introduced BisaFinder, an approach designed to uncover biased predictions in sentiment analysis systems using metamorphic testing. BisaFinder incorporates several key components, including the automatic generation of appropriate templates based on text fragments sourced from a large corpus. \textbf{Kiritchenko et al. \cite{kiritchenko2018examining}} introduced the \textbf{Equity Evaluation Corpus (EEC)}, which comprises 8,640 English sentences meticulously selected to uncover biases pertaining to specific races and genders. The primary objective of their research is to investigate 219 automatic sentiment analysis systems that participated in the SemEval-2018 Task 1 'Affect in Tweets' shared task, utilizing the EEC dataset. The researchers' analysis reveals that several of the sentiment analysis systems exhibit statistically significant biases. Specifically, these systems consistently generate slightly higher sentiment intensity predictions associated with one race or gender.

\subsection{Gender Bias in Language Modeling}

Lastly, gender bias can also occur in language modeling, where statistical and probabilistic techniques are employed to estimate the probability of a specific word string appearing in a sentence. These language models analyze vast amounts of text data to establish the basis for their word predictions. However, biases may inadvertently seep into these models, such as the incorrect association between the genders of men and women and gender-neutral roles. Some examples of language models include RNNs and LSTMs.

\begin{table*}[htbp]
	\centering
	\caption{Examples of Gender Bias in different tasks}
	\label{tasks}
	\resizebox{\linewidth}{!}{
	\begin{tabular}{|c|p{55mm}|c|c|c|p{45mm}|}
	\hline
	\textbf{Task} &  
    \centering \textbf{ Example of Representation Bias in the Context of Gender} & 
    \centering \textbf{A1} & 
    \centering \textbf{A2} & \centering \textbf{A3} &
    \centering \textbf{References} \tabularnewline 
\hline
Word Embeddings &
\centering Automatic Generation of analogies like "{\bf \textit{Man: Woman :: Programmer: Homemaker}}" &
\centering $\textcolor{green}{\checkmark}$ &
\centering $\textcolor{green}{\checkmark}$ &
\centering $\textcolor{green}{\checkmark}$ &
\centering Amazon \cite{WinNT}, Caliskan et al. \cite{caliskan2022gender}\tabularnewline
\hline

Machine Translation &
\centering "He is a nurse. She is a doctor." to Hungarian and back to English results in "She is a nurse. He is a doctor." &
\centering $\textcolor{red}{\times}$ &
\centering $\textcolor{green}{\checkmark}$ &
\centering $\textcolor{red}{\times}$ &
\centering Savoldi et al. \cite{10.1162/tacl_a_00401}\tabularnewline
\hline

Caption Generation &
\centering An image captioning model incorrectly predicts the agent to be male &
\centering $\textcolor{red}{\times}$ &
\centering $\textcolor{green}{\checkmark}$ &
\centering $\textcolor{red}{\times}$ &
\centering Hendricks et al. \cite{hendricks2018women}, Hirota et al. \cite{hirota2022quantifying} \tabularnewline
\hline

Sentiment Analysis &
\centering Sentiment Analysis Systems rank sentences containing female noun phrases to be indicative of anger more often than sentences containing Male noun phrases &
\centering $\textcolor{red}{\times}$ &
\centering $\textcolor{green}{\checkmark}$ &
\centering $\textcolor{red}{\times}$ &
\centering Park et al. \cite{park2018reducing}, Asyrofi et al. \cite{9653830}, Kiritchenko et al. \cite{kiritchenko2018examining} \tabularnewline
\hline

Language Model &
\centering ``He is a doctor" has a higher conditional likelihood than ``She is a doctor"&
\centering $\textcolor{red}{\times}$ &
\centering $\textcolor{green}{\checkmark}$ &
\centering $\textcolor{green}{\checkmark}$ &
\centering Lu et al. \cite{lu2020gender}, Bordia et al. \cite{bordia2019identifying} \tabularnewline
\hline

	\end{tabular}}
\end{table*}

\section{Representation of Gender Bias}

The issue of gender bias in various NLP tasks is a matter of significant concern, and it is essential to categorize and summarize the different forms of bias that can arise. Researchers have identified three primary categories of bias: Denigration (A1), Stereotyping (A2), and Under-representation (A3). \textbf{Denigration} involves the usage of racial, ethnic, or religious slurs, which can often be observed as a prevalent method of cyberbullying. This type of bias manifests as derogatory language aimed at demeaning specific groups. \textbf{Stereotyping}, on the other hand, refers to individuals' cognitive representation of a particular social group. In NLP tasks, stereotyping can manifest through word embeddings and large-scale language models, where certain biases are learned and perpetuated in the resulting representations. \textbf{Under-representation} pertains to the absence of identifiable group members from representative bodies and well-being indicators in proportion to their population's size. This type of bias highlights the disparities in visibility and inclusivity within various domains. Researchers have extensively studied these categories of bias to better understand their nature and mitigate their negative impact on society. By recognizing and categorizing different types of bias, researchers and practitioners can develop effective strategies and techniques to address gender bias in NLP. The above-mentioned NLP tasks can be represented and examined using the terms specified in Table \ref{tasks}. This categorization facilitates a clearer understanding of how gender bias can manifest in different NLP applications, thereby guiding efforts to combat and mitigate bias effectively. By comprehensively analyzing and addressing the various types of bias, researchers and practitioners can contribute to the development of more equitable and fair NLP systems and applications, promoting inclusivity and fostering unbiased linguistic representations.

\section{Why does Gender Bias Occur?}
According to a study by \textbf{Hovy et al. \cite{hovy2021five}}, there can be four primary sources of bias in NLP. These include bias from data, annotations, input representations, and models, which can be summarized in Table \ref{tab: biassource}. 

\subsection{Data Bias}
Biases in language training data have been extensively documented and recognized as a significant challenge in NLP systems. The prevalent sources of training data for NLP models often come from well-known news outlets, which tend to represent a limited demographic profile. These sources predominantly reflect the perspectives of individuals who are white, upper-middle-class, middle-aged, and educated \cite{garimella-etal-2019-womens}. Consequently, NLP models trained on such data inherit and perpetuate the demographic bias in the training samples, leading to biased predictions and outputs. The biased behavior of NLP models stems from learning patterns and associations from the training data. If the data is skewed towards a specific demographic, the models will reflect and reinforce that bias. For example, a model trained on news articles that predominantly feature male politicians may associate leadership roles with masculinity, leading to biased predictions in gender-related tasks. Furthermore, many syntactic tools, such as taggers and parsers, rely on outdated newswire data from the 1980s and 1990s. These tools can inadvertently reinforce biases by assuming that everyone speaks and writes in a manner similar to journalists from that era. This temporal bias can result in models failing to understand or accurately represent the language used by diverse demographic groups, further perpetuating linguistic inequality. Given these challenges, it is crucial to consider the demographic representation within the chosen text data collection. Even seemingly neutral or unbiased datasets can carry latent biases due to the inherent demographic signals embedded in language itself. Therefore, researchers and practitioners must be mindful of their training data's limitations and potential biases. By carefully considering the demographic groupings represented in the data and actively working towards mitigating biases, NLP researchers and practitioners can strive to develop more fair and equitable systems. This entails promoting diversity and inclusivity in both the training data and the development process, leading to NLP models that better understand and respect the rich linguistic variations present in society.

\subsection{Annotation Bias}
\textbf{Annotation bias}, also known as \textbf{label bias}, is a phenomenon in NLP where human annotators inadvertently introduce biases into labeled data used for training and evaluating NLP models \cite{lingren2014evaluating}. This bias can significantly impact the performance and fairness of these models. There are several factors that can contribute to annotation bias. Annotators may hold certain assumptions or stereotypes about the language or task they are working on, which can influence their labeling decisions. For example, they might have preconceived notions about the sentiment of a particular text or the gender associated with certain occupations. Furthermore, annotators can be influenced by the data they are annotating, especially if the data itself contains biases, as mentioned in the previous section. If the training data exhibits imbalances or reflects societal biases, annotators may inadvertently align their labels with those biases, reinforcing them in the labeled dataset. Addressing annotation bias is essential for developing NLP models that are fair, robust, and respectful of diverse perspectives. It is essential to be aware of its presence and take proactive steps during the annotation process to mitigate annotation bias. Providing clear guidelines to annotators that explicitly address potential biases and instruct them to label based on the content rather than their assumptions or stereotypes can help reduce bias. Regular training and discussions with annotators can promote awareness and sensitivity toward bias. Using multiple annotators and measuring inter-annotator agreement can help identify and address inconsistencies or biases in the labeled data. Adjudication or consensus mechanisms can resolve disagreements among annotators, ensuring a more balanced and unbiased representation of the data. Furthermore, ongoing efforts are being made to develop methods that explicitly model and mitigate annotation bias while training NLP models. By explicitly accounting for the biases introduced during annotation, reducing their impact on the model's predictions and improving fairness is possible. By acknowledging and actively mitigating annotation bias, the NLP community can strive towards more accurate and equitable NLP systems that better reflect natural language's varied nuances and characteristics.

\begin{figure*}[t]
    \centering
    \includegraphics[width = \linewidth]{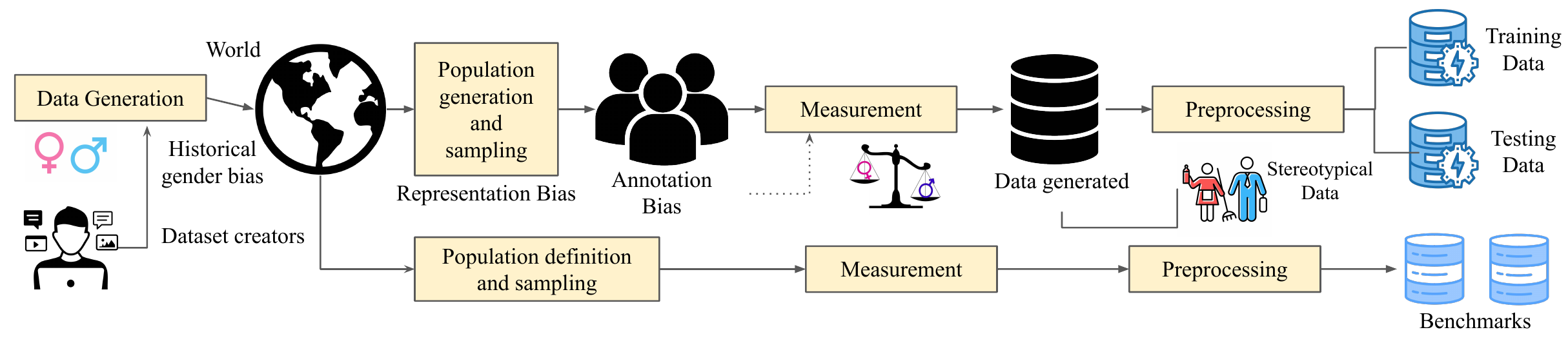}
    \caption{Gender Bias induced from Data Generation}
    \label{fig:DataBias}
\end{figure*}

\subsection{Bias from Input Representations}
Bias from input representations in NLP refers to the introduction of bias into the input data used for training and evaluating NLP models. These biases, often referred to as semantic biases, can arise from various sources and have significant implications for the fairness and accuracy of NLP systems. One common source of bias is the lack of representativeness in the input data. If the training data does not adequately reflect the diversity of the population or the specific task it aims to address, biases can emerge. For example, if the training data predominantly represents a particular demographic group or region, the model's predictions may be skewed towards that group's perspectives and experiences, leading to biased outputs. Another factor contributing to bias in input representations is the pre-processing of data. Pre-processing steps, such as text normalization, tokenization, or stemming, can inadvertently introduce biases. For example, certain linguistic variations or expressions commonly used by specific demographic groups may be overlooked or normalized, resulting in a loss of cultural nuances and potential bias in the representations.

Studies have demonstrated that word embeddings can detect racial and gender biases, even in well-labeled and balanced datasets. These biases can arise due to societal prejudices reflected in the training data, leading to biased predictions and outputs from NLP models. Contextual representations learned by large pre-trained language models, such as BERT and GPT, are also susceptible to biases. These models are typically trained on vast amounts of internet text, including societal biases in online content. Consequently, these models can replicate and perpetuate biases, often mirroring societal biases. Numerous studies have documented and quantified biases in NLP models and their input representations, highlighting the importance of addressing these issues. Recognizing and understanding the biases generated during the data generation process is a critical step towards mitigating them. Addressing bias in input representations requires a multi-faceted approach. It involves diversifying training data sources to ensure the representation of various demographic groups and perspectives. Regularly evaluating and auditing the models for biases and developing debiasing techniques are crucial to mitigating these biases. The above-mentioned biases are collectively represented as the biases generated in the process of data generation, which can be depicted in Fig. \ref{fig:DataBias}

\subsection{Model Bias}
\begin{figure}[htbp]
    \centering
    \includegraphics[width = \linewidth]{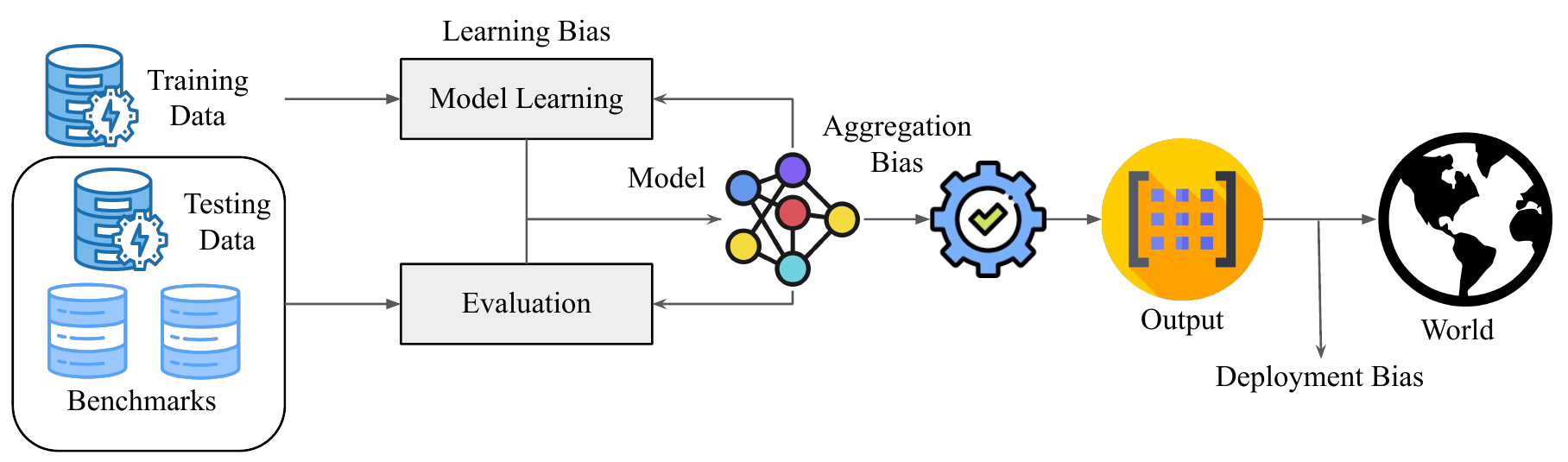}
    \caption{Gender Bias Induced from Model Building}
    \label{fig:modelBias}
\end{figure}

Languages are dynamic and constantly evolving, making capturing their complexity and nuances challenging even with a large dataset. Using a small subset of data can only provide a limited and temporary snapshot of language, which is why relying solely on ``better" training data is not a comprehensive solution to address bias in NLP models. Furthermore, machine learning models tend to amplify the behaviors and patterns they are exposed to, including biases present in the training data. Studies such as \cite{kiritchenko-mohammad-2018-examining, hovy2020you} have explored the compounding effect of bias in newer models, highlighting the phenomenon known as bias overamplification. This refers to the tendency of machine learning models to disproportionately amplify and reinforce biases rather than mitigate them. One contributing factor to bias overamplification in language models is the choice of loss objective used during training. Often, these objectives prioritize improving the model's prediction accuracy, which can incentivize the model to exploit spurious correlations or irregularities in the training data. As a result, the model may rely on certain discriminatory features, such as gender or race, to achieve higher accuracy, even if those features are irrelevant to the task. This behavior is challenging to detect until a consistent pattern of bias is identified and examined. Addressing bias overamplification requires a more nuanced and comprehensive approach beyond improving the training data. It involves reevaluating the loss objectives and training methods to incorporate fairness and mitigate biases. Researchers and practitioners are actively exploring techniques to promote fairness and reduce bias in NLP models, such as incorporating fairness constraints, developing debiasing algorithms, or redefining evaluation metrics to account for bias. Moreover, addressing bias overamplification requires collaboration and engagement with diverse stakeholders, including linguists, ethicists, and impacted communities. These collaborations can help in uncovering and understand the complexities of bias in language models and develop more holistic approaches to mitigate bias overamplification. Fig. \ref{fig:modelBias} gives a clear pictorial representation of model bias. 

\begin{table*}[t]
	\centering
	\caption{Tabulated overview of Bias Sources in NLP}
	\label{tab: biassource}
	\resizebox{\linewidth}{!}{
	\begin{tabular}{|p{20mm}|p{55mm}|p{60mm}|p{30mm}|}
	\hline
	\centering \textbf{Source of Bias} &  
    \centering \textbf{Description} &
    \centering \textbf{Key Characteristics} &
    \centering \textbf{References} \tabularnewline 
    \hline

    \centering Data Bias &
    \centering Bias present in the training data used for NLP models &
    \centering Limited demographic representation, skewed patterns, and associations, temporal bias from outdated sources &
    \centering Garimella et al. \cite{garimella-etal-2019-womens} \tabularnewline 
    \hline

    \centering Annotation Bias &
    \centering Bias introduced during the process of data labeling and annotation &
    \centering Assumptions and stereotypes of annotators, alignment with biases in training data, inconsistencies and disagreements among annotators  &
    \centering Lingren et al. \cite{lingren2014evaluating}\tabularnewline 
    \hline 

    \centering Bias from Input Representations &
    \centering Bias arising from the representativeness and pre-processing of input data  &
    \centering Lack of diversity in input data, loss of cultural nuances and biases introduced during pre-processing steps, biases in word embeddings and contextual representations &
    \centering Peng et al. \cite{peng2019you}\tabularnewline 
    \hline 
    
    \centering Model Bias &
    \centering Bias amplified and reinforced by the NLP models themselves &
    \centering Bias overamplification, reliance on discriminatory features, choice of loss objectives favoring accuracy over fairness &
    \centering Kiritchenko et al. \cite{kiritchenko-mohammad-2018-examining}, Hovy et al. \cite{hovy2020you} \tabularnewline 
    \hline

	\end{tabular}}
\end{table*}

\section{Transformers and Gender Bias}
In recent years, the popularity of Language Modelling has significantly increased, mainly due to the development of transformers such as BERT, GPT-2, and XLM \cite{wolf2019huggingface, gillioz2020overview, bracsoveanu2020visualizing}. These deep learning models employ a self-attention process that allows them to weigh the importance of different input data components differently \cite{vaswani2017attention}. Additionally, transformer-based models can be fine-tuned for a specific downstream task, making them highly versatile. Fine-tuning requires much less data than training a language model from scratch, making these models highly efficient. However, despite their efficiency, gender bias has been observed in transformers, indicating that the problem of bias in NLP is still prevalent. Evidence of gender bias in transformer models like GPT2 \cite{budzianowski2019hello}, GPT3 \cite{floridi2020gpt}, RoBERTa \cite{liu2019roberta}, and DeBERTa \cite{he2020deberta} was proven by researchers across the globe through a series of experiments. To identify occupational gender bias in GPT-2, examine how the prejudice evolves with various model sizes, and contrast this bias with bias in our culture, \textbf{Bolukbasi et al. \cite{bolukbasi2016man}} conducted several tests. The experiments have shown that occupations became more gender-neutral as the number of trained parameters increased. According to societal statistics and all four GPT-2 models, there is a trend toward increased male bias as job salaries rise. That is, the more senior the job and the bigger its monetary compensation, the more likely it is that a man is holding that position, according to GPT-2.

Similarly, a study conducted by \textbf{Brown et al. \cite{brown2020language}} has also proven that GPT-3 contains racial and gender bias. On performing the occupational experiment, results have shown that GPT-3 found that 83\% of the 388 evaluated vocations were more likely to be linked to a male identity. Also, higher-level occupations with a predominance of men included banker and professor emeritus. Another well-known AI chatbot, ChatGPT \cite{lund2023chatting}, has been accused of gender bias \cite{borji2023categorical, ortega2023linguistic}. Kieran Snyder, a co-founder of Textio, points out that ChatGPT can start incorporating gendered presumptions into feedback that is otherwise so generic with very little effort. The feedback that is of a high caliber concentrates on an individual's work rather than their personality. It offers precise, pertinent examples. It is concise, pertinent, and straightforward. Men were noticeably more likely than women to be labeled as ambitious and confident in Textio's groundbreaking analysis of performance evaluation received by 25,000+ workers at 250+ organizations, while women were more likely to be regarded as collaborative, helpful, and outspoken. They represent the precise bias trends that appear in ChatGPT's written comments. Researchers are currently working on extending transformers to other types of input, particularly visual inputs, because they have shown to be so helpful in NLP. The attention processes that the transformer employs are responsible for its performance in each of these domains. Models can selectively focus on a small number of pertinent parameters while ignoring others thanks to attention processes. Also, a transformer considers (possibly) all data at once. Therefore, the influence is all against all rather than one against the next, whereas in standard neural networks, the processing of one item influences the processing through recursion and changes the way the following one is processed.

Contrary to popular belief, this strategy really uses less computing power. When more data spaces are processed and more become available, the use of transformers will spread and result in acceleration. Several of the main similarities between how people process information and learn across a wide range of tasks appear to have been incorporated into the transformer \cite{khan2022transformers}. The similarities to human learning provide a positive outlook on the transformer's potential in the future. In other AI disciplines, current research points to a variety of novel applications for transformers, such as teaching robots to recognize human body movements \cite{jangir2022look}, teaching computers to understand emotions in speech, and identifying stress levels in electrocardiograms. Due to their versatility, transformers can be considered the future of AI \cite{han2021pre}, and it is highly considered essential to identify and mitigate gender bias in these models. The following are the major highlights of the survey presented in this research.
\vspace{-7.5px}
\begin{itemize}
    \item In our research, we emphasize various scenarios of gender bias, identify sources of bias, and articulate the linguistic consequences of gender bias in the latest Transformer models. We have categorized these consequences in a clear and concise manner, enabling a comprehensive understanding of the issue.  
    \item Furthermore, we provide a detailed overview of cutting-edge techniques utilized to detect and alleviate gender bias in Transformer Models. By thoroughly examining and analyzing each proposed methodology from diverse researchers worldwide, we also discuss the experiments conducted and datasets used to obtain the results. 
    \item Our survey of gender bias in Transformer models stands out as the most comprehensive to date, as we not only highlight the shortcomings of the proposed solutions mentioned above and examine future development prospects but also establish rules and ethical guidelines for measuring gender bias.
\end{itemize}

\section{Examining Gender Bias Detection Techniques in Transformer Models}
When evaluating gender bias in transformer models, it's important to use a combination of metrics to comprehensively understand the model's performance. While individual metrics can provide insights into specific aspects of bias, they may not capture the full extent of gender bias present in the model. As a result, combining multiple metrics can provide a more accurate assessment of gender bias. There exist a variety of metrics that can be employed to evaluate gender bias in transformer models whose summary can be illustrated in Table \ref{tab: biasdetectionmethods}

\subsection{WEAT score}
The \textbf{WEAT (Word Embedding Association Test) score} measures the degree of association between two sets of words based on their embedding vectors in a language model, such as a transformer model. Specifically, it measures the degree of association between a set of target words and a set of attribute words. In the context of gender bias in transformer models, the WEAT score can be used to evaluate whether the model is exhibiting biased associations between gendered words and certain attributes (such as career vs. family). A higher WEAT score indicates a stronger association between the target and attribute sets, which could suggest the presence of bias in the model. The WEAT score is based on the concept of "cosine similarity", which is a measure of the similarity between two vectors. In the context of word embeddings, cosine similarity is used to measure the similarity between the vectors representing two words. To calculate the WEAT score, four sets of words are first defined: \textit{Set A} consisting of words that are stereotypically associated with one gender \textit{"man," "male," "he," "brother"}, \textit{Set B}, consisting of words that are stereotypically associated with the other gender \textit{"woman," "female," "she," "sister"}. \textit{Set X} consists of words that are associated with a specific attribute \textit{"career," "professional," "executive"} and \textit{Set Y} consists of words that are associated with a contrasting attribute \textit{"family," "home," "parent"}.
The WEAT score is then calculated as the difference between the average cosine similarities of words in sets $A$ and $X$, and the average cosine similarities of words in sets $B$ and $X$, normalized by the standard deviation of the cosine similarities of all words in the sets:
\begin{equation}
\mathrm{W}(X,Y,A,B) = \\ \frac{\sum_{w \in X} \mathrm{cos}(w, A) - \sum_{w \in X} \mathrm{cos}(w, B)}{\sqrt{\sum_{w \in X} (\mathrm{cos}(w, A) - \mathrm{cos}(w, B))^2}}
\end{equation}
where $cos(w, S)$ is the cosine similarity between the word $w$ and the average of the vectors in set $S$. A higher absolute value of the WEAT score indicates a stronger association between the attributes $X$ and $Y$ and the gendered words in sets $A$ and $B$. \textbf{A WEAT score of 0 indicates that there is no difference in the association between the two sets of gendered words and the attributes $X$ and $Y$}. This metric was illustrated by \textbf{Silva et al. \cite{silva2021towards}} with the investigation of gender bias in transformers like GPT-2 and XLNet by employing WEAT as one of the methodologies used to detect gender bias. According to the findings, RoBERTa is one of the most consistently biased transformers among other models.

\subsection{Equalized Odds}
\textbf{Equalized Odds} is a fairness metric that can be used to detect gender bias in transformer models \cite{pmlr-v108-awasthi20a, garg2020fairness}. The metric measures the degree to which a model's predictions are equal across different demographic groups, such as males and females. In the context of gender bias, the metric can be used to assess whether the model is making equally accurate predictions for male and female inputs. The Equalized Odds metric is calculated by comparing the \textbf{True Positive Rates (TPRs)} and \textbf{False Positive Rates (FPRs)} across different demographic groups for a given prediction task. In the context of gender bias, this typically involves comparing the TPRs and FPRs for male and female inputs. The TPR measures the proportion of positive cases that are correctly identified by the model, while the FPR measures the proportion of negative cases that are incorrectly identified as positive by the model. By comparing the TPRs and FPRs across demographic groups, the Equalized Odds metric can provide a measure of how fairly the model is making predictions for different groups. A lower Equalized Odds score indicates a lower level of gender bias in the model's predictions, as it suggests that the model is making equally accurate predictions for males and females. The TPRs and FPRs for each demographic group are first computed to calculate the Equalized Odds metric. The metric is then calculated as the maximum difference in TPRs or FPRs across demographic groups. Specifically, the metric is the maximum of the absolute differences in TPRs or FPRs for any given threshold:

\begin{equation}
\mathrm{Equalized\ Odds} = \max_{t \in [0, 1]} \left| \mathrm{TPR}{\mathrm{male}}(t) - \mathrm{TPR}{\mathrm{female}}(t) \right| 
\end{equation}

\begin{equation}
\mathrm{Equalized\ Odds} = \max_{t \in [0, 1]} \left| \mathrm{FPR}{\mathrm{male}}(t) - \mathrm{FPR}{\mathrm{female}}(t) \right|
\end{equation}

\subsection{Counterfactual evaluation}
\textbf{Counterfactual evaluation} is a technique used to measure gender bias in transformer models by assessing the impact of gender swapping on model performance. The method involves modifying the gender of words in a dataset and observing the effect on the model's accuracy and other performance metrics. To apply this method, a dataset is first split into a training set and a test set. Then, the gender of the words in the test set is modified by swapping gendered pronouns or replacing gendered words with gender-neutral alternatives. For example, the word "he" could be replaced with "they", or the name "John" could be replaced with "Alex". The modified test set is then used to evaluate the model's performance, comparing the results to those obtained from the original test set. The difference in accuracy and other performance metrics between the two sets is used to calculate the degree of gender bias present in the model. Let the original test set be denoted by X and its associated labels by Y. Let the modified test set, where gendered words have been replaced with gender-neutral alternatives, be denoted by X', with associated labels Y'. Let the model's predicted labels on X and X' be denoted by $Y_{hat}$ and $Y_{hat}'$, respectively. The first step is to calculate the baseline performance of the model on the original test set:
\begin{equation}
    acc_{orig} = \frac{1}{|X|} \sum_{i=1}^{|X|} [Y_i = Y_{hat,i}]
\end{equation}
where $acc_{orig}$ is the accuracy of the model on the original test set, $|X|$ is the number of examples in the test set, $Y_i$ is the true label of example $i$, and $Y_{hat,i}$ is the predicted label of example i. Next, the performance of the model on the modified test set is calculated:
\begin{equation}
    acc_{mod} = \frac{1}{|X'|} \sum_{i=1}^{|X'|} [Y_i' = Y_{hat,i}']
\end{equation}
where $acc_{mod}$ is the accuracy of the model on the modified test set, $|X'|$ is the number of examples in the modified test set, $Y_i'$ is the true label of the gender-swapped example i, and $Y_{hat,i}'$ is the predicted label of the gender-swapped example I. The degree of gender bias in the model can be calculated as the difference between the two accuracies:
\begin{equation}
    bias_{gender} = acc_{orig} - acc_{mod}
\end{equation}
A positive value of $bias_{gender}$ indicates that the model is biased towards the original gender, while a negative value indicates a bias towards the opposite gender. A value of zero indicates no gender bias in the model. The counterfactual evaluation method provides a way to measure gender bias in transformer models without relying on external benchmarks or human annotations. By simulating the impact of gender-swapping on model performance, the technique can identify cases where the model relies on gendered cues to make predictions rather than on other relevant information in the text. One limitation of the counterfactual evaluation method is that it only measures the impact of gender on performance in a binary sense (i.e., male versus female). Other factors, such as race, ethnicity, or sexuality, may also contribute to bias in the model but are not captured by this method. Additionally, the technique assumes that gender-neutral replacements are available for all gendered words in the dataset, which may not always be the case. Overall, counterfactual evaluation is a valuable method for detecting and quantifying gender bias in transformer models, providing a complement to other techniques such as WEAT and Equalized Odds.

\subsection{BLEU Score}
BLEU (Bilingual Evaluation Understudy) score is a metric commonly used to evaluate the performance of machine translation systems. However, it can also be used to estimate gender bias in transformer models. The basic idea is to use gender-swapped versions of sentences as inputs and compare the similarity of the model's outputs with the original sentences. To apply this method, a dataset is first split into a training set and a test set. Then, gender-swapped versions of sentences in the test set are created, where the gendered words are replaced with their gender-neutral counterparts. For example, the sentence "He went to the store" would be swapped to "They went to the store". The gender-swapped sentences are then fed into the model, and the output is compared to the original sentence using the BLEU score. The BLEU score measures the degree of overlap between the n-grams (subsequences of n words) in the model's output and the original sentence, with higher scores indicating greater similarity. The BLEU score is calculated using the following equation:
\begin{equation}
    BLEU = BP \times exp(\frac{1}{n} \sum_{i=1}^n log p_i)
\end{equation}
where BP is the brevity penalty, which adjusts the score based on the length of the output sentence compared to the original sentence, and $p_i$ is the precision score for n-grams of length i. The precision score is calculated as the number of n-grams in the model's output that appear in the original sentence divided by the total number of n-grams in the model's output. To estimate gender bias using the BLEU score, the average score for male and female gender-swapped sentences is calculated and compared. If the model consistently produces higher BLEU scores for male gender-swapped sentences than for female ones, this indicates a bias towards male language. Conversely, if the model consistently produces higher BLEU scores for female gender-swapped sentences than for male ones, this indicates a bias towards female language. One advantage of the BLEU score is that it is a widely used and standardized metric, making comparing results across different studies easier. However, it has some limitations, as it only measures surface-level similarity between the model's output and the original sentence and does not capture more subtle forms of gender bias, such as implicit associations or stereotypes.

\begin{table*}[htbp]
	\centering
	\caption{Tabulated overview of Gender Bias Detection Techniques in Transformer Models}
	\label{tab: biasdetectionmethods}
	\resizebox{\linewidth}{!}{
	\begin{tabular}{|p{20mm}|p{75mm}|p{80mm}|}
	\hline
	\centering \textbf{Technique} &  
    \centering \textbf{Methodology} &
    \centering \textbf{Limitations} \tabularnewline 
    \hline

    \centering WEAT score &
    \raggedright 
    1. Measures the degree of association between gendered words and certain attributes using word embedding vectors. \\
    2. Higher score indicates stronger association and potential bias. &
    \raggedright
    1. Word Embedding limitations and a limited set of words may not capture the full range of gendered associations. \\
    2. Lack of Contextual Information and Subjectivity of Attribute Definitions. \tabularnewline 
    \hline

    \centering Equalized Odds &
    \raggedright 
    1. Evaluates whether a model's predictions are equal across different demographic groups. \\
    2. Specifically, compares true positive rates and false positive rates for males and females. &
    \raggedright
    1. Binary Classification Focus may not directly apply to more nuanced gender categories or non-binary gender identities.  \\
    2. Assumption of Independence and Trade-offs and Information Loss resulting in reduced overall accuracy \tabularnewline 
    \hline

    \centering Counterfactual evaluation&
    \raggedright 
    1. Measures gender bias by assessing the impact of gender swapping on model performance. \\
    2. Compares accuracy on original and gender-swapped test sets to quantify the degree of bias. &
    \raggedright
    1. Availability of Counterfactual Data and the definition of Counterfactual Scenarios; selecting unrealistic or arbitrary counterfactual scenarios can lead to misleading assessments of bias \\
    2. Limited Scope of Counterfactuals and Causal Inference Challenges exist in this methodology \tabularnewline 
    \hline

    \centering BLEU Score &
    \raggedright 
    1. Calculates the similarity between the model's output and the original sentence using gender-swapped sentence versions. \\
    2. Used to estimate gender bias based on score differences.&
    \raggedright
    1. BLEU primarily measures the lexical overlap between the generated and reference translations. It does not capture language's semantic or contextual aspects, including gender bias.\\
    2.  BLEU does not explicitly consider gender-related biases in the translations. It treats all words and phrases equally without accounting for potential stereotypes, imbalances, or unequal treatment based on gender. \tabularnewline 
    \hline

    \centering Stereoset &
    \raggedright 
    1. Utilizes crowd-sourcing to evaluate stereotyped evaluations made by MLMs on gender and career. \\ 
    2. Provides bias scores on a scale of 0 to 100, with lower scores indicating less bias.&
    \raggedright
    1. Limited to MLMs as Stereoset relies on predefined sets of stereotypical sentences, which may not cover the entire spectrum of possible biases. \\ 
    \tabularnewline 
    \hline

    \centering Attention Maps &
    \raggedright 
    1. Analyzes attention maps in transformer-based models to detect gender bias. \\ 
    2. Compares the relation degree between genders and occupations based on attention scores to identify bias-contributing modules.&
    \raggedright
    1. Attention scores themselves don't always have a direct and intuitive correspondence to the importance or significance of specific features or connections. Interpreting attention maps requires careful analysis and domain expertise.\\ 
    2. Transformer models often exhibit sparse attention, meaning that only a small subset of the input tokens receive significant attention weights. This sparsity can limit the granularity of the analysis and make it difficult to capture fine-grained biases. 
    \tabularnewline 
    \hline

	\end{tabular}}
\end{table*}

\subsection{Stereoset}
\textbf{Robinson et al. \cite{robinson2021assessing}} proposed a technique for identifying gender bias in both traditional and  Medical \textbf{Masked Language Models (MLMs)} such as SciBERT \cite{beltagy2019scibert} and BioClinicalBERT. Medical MLMs refer to language models that are specifically pre-trained on medical text. These models have the potential to improve the accuracy and speed of medical text analysis and provide new insights into clinical data. Unlike general-purpose MLMs, medical MLMs are pre-trained on a large corpus of medical text, including scientific publications, clinical notes, and electronic health records. This allows the models to capture the unique language and terminology used in the medical field and the specific contexts in which these terms are used. The proposed methodology, called StereoSet, uses a collection of 17,000 test words and crowd-sourcing to quantify stereotyped evaluations about gender and career made by MLMs. StereoSet evaluates the most likely word chosen by an MLM to fill in the blank in intra-sentence and inter-sentence Context Association Tests (CATs). Bias scores are given on a scale of 0 (strong bias) to 100 (extremely low or no bias), and BERT achieved a gender bias score of 63, while RoBERTa achieved a score of 73. On the other hand, the medical MLMs exhibited more bias in all categories than the general-purpose MLMs, except for SciBERT, which showed a better race bias score of 55 than BERT's 53. The medical MLMs also showed more gender and religious biases compared to the general-purpose MLMs. The evaluation of four medical MLMs for stereotyped assessments about race, gender, religion, and profession revealed lower performance compared to general-purpose MLMs. These medically-focused MLMs differ considerably in their training source data, which likely contributes to the differences in the ratings for stereotyped biases from the StereoSet tool. Overall, this study highlights the importance of considering and addressing biases in NLP systems, particularly those used in sensitive areas such as healthcare and science, where accurate and unbiased results are crucial.

\subsection{Attention Maps}
\textbf{Li et al. \cite{li2021detecting}} presented a novel gender bias detection method for transformer-based models by utilizing attention maps. The authors propose an intuitive gender bias judgment method by comparing the relation degree between genders and occupations based on attention scores. They also design a gender bias detector by modifying the attention module and inserting it into different positions of the model to present the internal gender bias flow. By scanning the entire Wikipedia, a BERT pre-training dataset, the authors draw a consistent gender bias conclusion. Their findings show that attention matrices, $W_{q}$ and $W_{k}$, introduce much more gender bias than other modules, including the embedding layer. The bias degree changes periodically inside the model, where the attention matrix $Q$, $K$, $V$, and the remaining part of the attention layer enhance gender bias, while the averaged attentions reduce the bias. This study is the first attempt to investigate gender bias inside transformer-based models, using BERT as an example, and provides insights into the mechanisms that contribute to gender bias in NLP models.

\section{Overview of Gender Bias Estimation Techniques in Transformer Models}
Over the years, researchers have proposed various methodologies for Gender Bias mitigation in transformer models. These methods include altering the training data, adjusting model architecture, and incorporating additional constraints during training. Each of these approaches has its own advantages and limitations. This section provides an overview of the different methodologies used for Gender Bias mitigation among various transformer models with its summary being portrayed in Table \ref{tab: methods}. We discuss the strengths and weaknesses of each approach and highlight the current state-of-the-art methods in this field. Through this discussion, we hope to provide a comprehensive understanding of Gender Bias mitigation in transformer models and its importance in creating fair and equitable NLP systems.

\subsection{Movement Pruning}
A technique used to inspect the gender bias in pre-trained language models using the attention layers was proposed by \textbf{Joniak et al. \cite{joniak2022gender}}, also known as movement pruning. When some weights are disabled or removed from a neural network, the process is known as pruning. The authors modified movement pruning, allowing one to select a low-bias subset of a given model or, more precisely, to identify the model weights whose removal causes an arbitrary debiasing objective to converge. The proposed strategy is innovative since it combines debiasing, weight freezing, and movement pruning. It also investigates if gender bias exists in a BERT model and suggests ways to improve an existing debiasing technique. Gender bias was used by the authors to demonstrate how to use their framework, and they discovered that the bias is primarily stored in the intermediate layers of BERT. The approach uses movement pruning to identify a subset that has less bias than the original model given a model and a debiasing aim. If a model produces more invariance, it may become faster and smaller while preserving its previous performance. They also noticed that there is a direct relationship between model bias and performance.

\subsection{Transfer Learning}
\begin{figure}[ht]
    \centering
    \includegraphics[width = \linewidth]{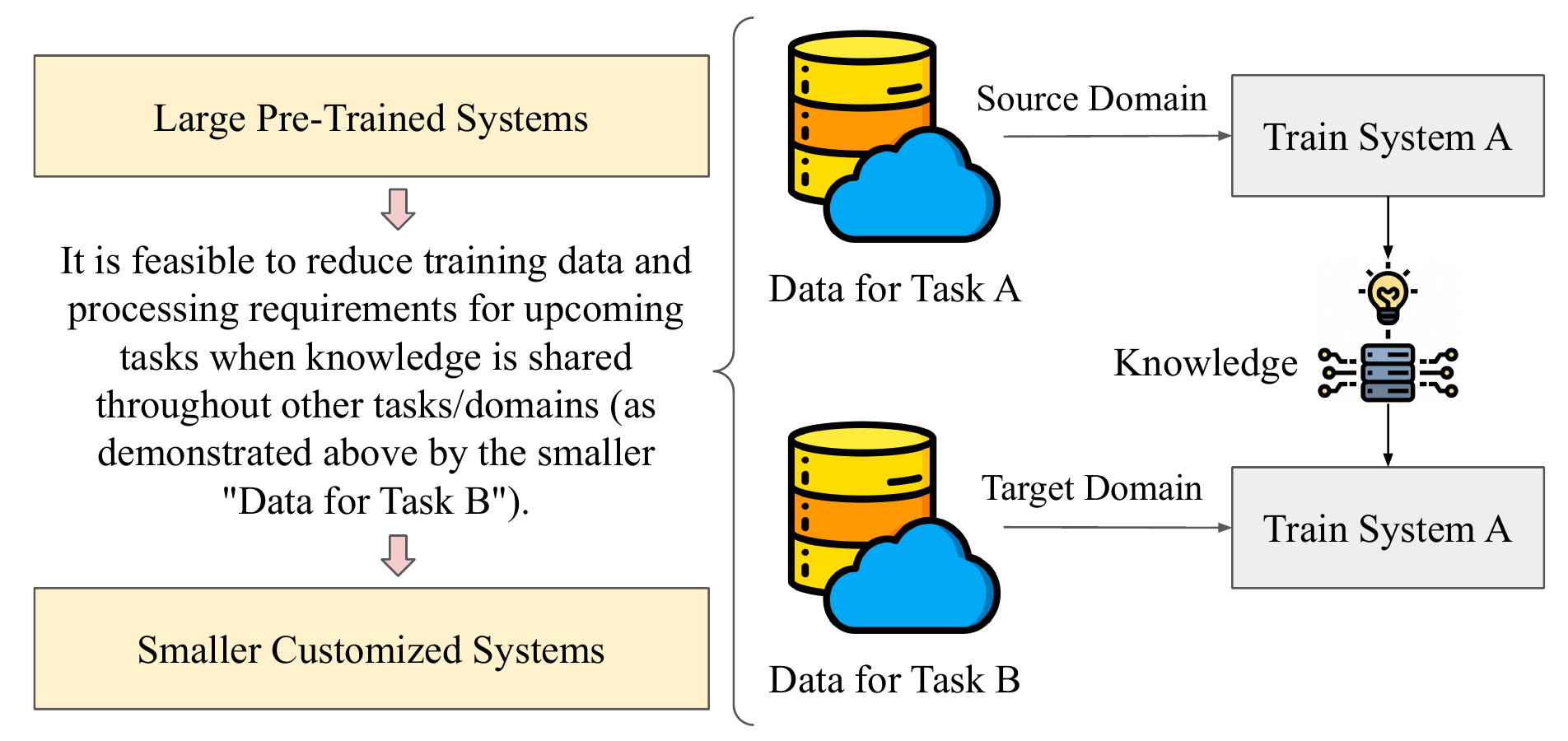}
    \caption{Transfer Learning Paradigm: Knowledge sharing between domains}
    \label{fig:TL}
\end{figure}

\textbf{Transfer learning} is a machine learning technique in which a model created for one task is utilized as the foundation for a model on another. In other words, it involves leveraging the knowledge learned from a pre-existing model to improve the performance of a new model on a related or different task. The main advantage of transfer learning is that it can significantly reduce the amount of training data and time required to achieve high performance on a new task. In transfer learning, the pre-existing model is typically a large and complex neural network that has been trained on a large dataset. The idea is to use this model as a starting point and then fine-tune it on the new task by updating the weights of some of the layers or adding new layers to the model. This way, the model can learn task-specific features while retaining the general knowledge learned from the pre-existing model. Mathematically, transfer learning can be represented as follows. Let $D_{s}$ and $D_{t}$ be the source and target domains, respectively, where $D_{s}$ is the domain on which the pre-existing model is trained and $D_{t}$ is the domain of the new task. Let $f_{s}$ be the pre-existing model and $f_{t}$ be the new model. The goal of transfer learning is to learn a mapping $h: D_{s} \rightarrow D_{t}$ that transfers knowledge from $f_{s}$ to $f_{t}$. The conceptual overview of transfer learning can be depicted in Fig. \ref{fig:TL}

\textbf{Bao et al. \cite{bao2019transfer}} investigated transfer learning from pre-trained models to improve task performance with little data. It has been demonstrated that the majority of current representative coreference systems suffered on the \textbf{GAP dataset} \cite{beamer2015gap}, performing only mediocrely overall and with significant gender differences in performance. These coreference systems' uneven training datasets or the systems' architecture may be to condemn for this. To enhance the performance of tasks requiring less data, the authors of this work investigated transfer learning from trained models. Furthermore, using data from the \textbf{Caliskan dataset} \cite{wolfe2021low}, a statistical experiment was conducted to examine gender bias in word and sentence-level embeddings.  Several efficient ways to reuse pre-trained BERT knowledge in this shared work are offered and compared. The resulting system outperforms off-the-shelf resolvers significantly, with balanced prediction performance for the two genders.

\subsection{Casual Mediation Analysis}
\begin{figure}[ht]
    \centering
    \includegraphics[width = \linewidth]{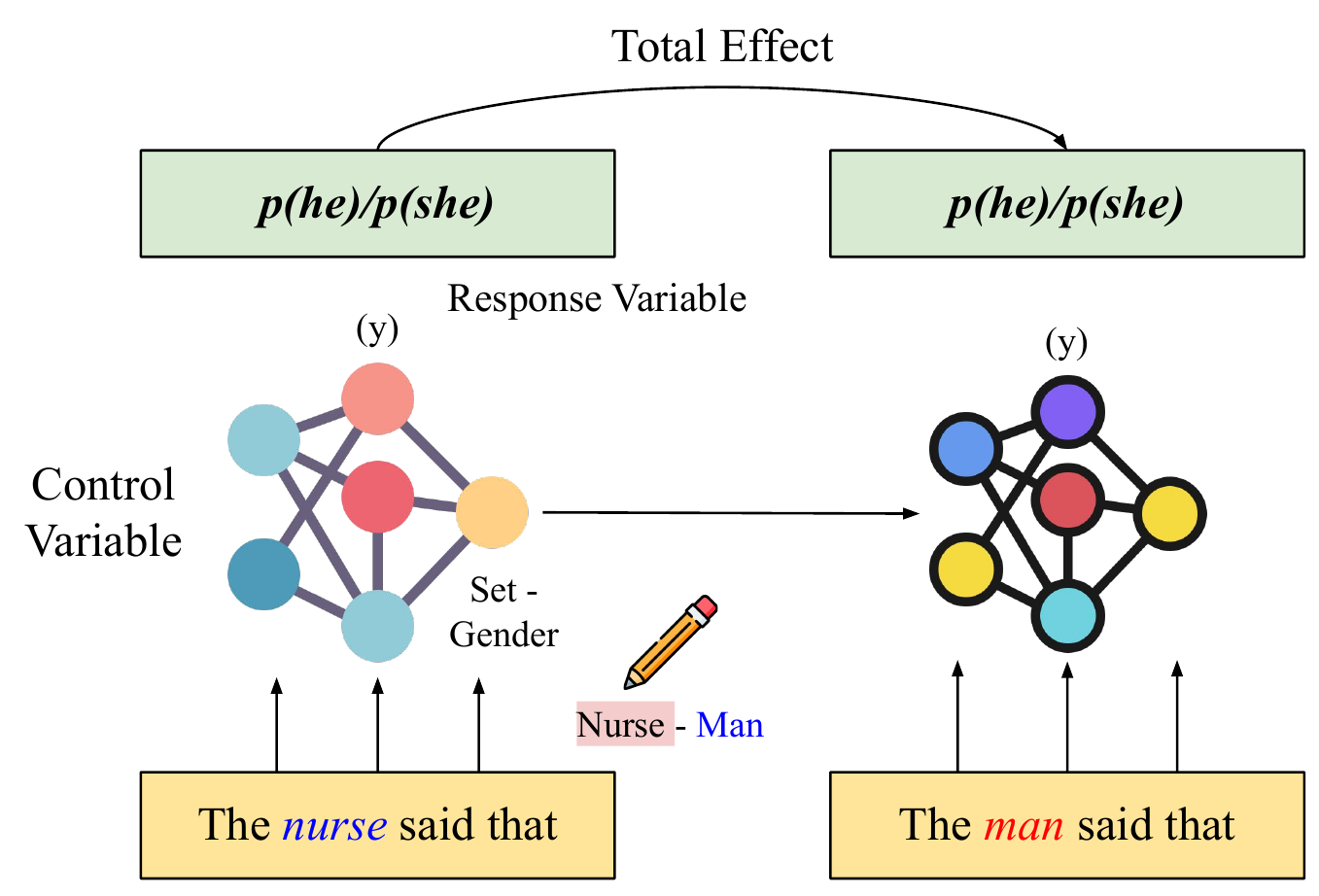}
    \caption{Casual Mediation Analysis}
    \label{fig:CMA}
\end{figure}

The goal of \textbf{Causal Mediation Analysis}, also known as CMA, is to determine the extent to which intermediary factors mediate a treatment effect. CMA separates a treatment's overall impact into its direct and indirect effects. The mediator, which is defined as the intermediate variable in the casual path, transmits the indirect effect to the result. The mediation package is made to execute CMA with the sequential ignorability assumption. Using the above concept, \textbf{Vig et al. \cite{vig2020investigating}} proposed a method to investigate the mechanisms that allow information to move from input to output via numerous model components. The authors described a gender-specific anti-stereotypical intervention set-gender that transforms the profession \textbf{nurse} into \textbf{man}. The total effect is denoted by the change in the response variable and the methodology is illustrated in Fig. \ref{fig:CMA}. They categorized the gender bias effects as sparse, concentrated in a limited area of the network, synergistic, enhanced or suppressed by different components, and decomposable into effects flowing directly from the input and indirectly through the mediators. Using three datasets designed to assess a model's susceptibility to gender bias, the authors investigated the function of individual neurons and attention heads in influencing gender bias. This mediation study demonstrates that the effects of gender prejudice are focused on specific model components that may exhibit highly specialized behavior. The transformers involved in this study are \textbf{GPT2, XLNet, RoBERTa,} and {\bf DistilBERT}.

\subsection{MLP Regression}
\textbf{Bharadwaj et al. \cite{bhardwaj2021investigating}} addressed that CLMs are prone to learning the dataset's inherent gender bias. As a result, changing gender words—such as switching "he" for "her" or using gender-neutral words—can lead to dramatically different predictions from downstream NLP models. They concentrated on a well-known CLM, {\bf BERT}. For several NLP tasks, they trained a basic regressor using BERT's word embeddings \cite{reimers2019sentence} and then evaluated the gender bias in regressors using an equality evaluation corpus. Ideally, depending on the design, the models should not accept input that contains gender-specific information. The findings show that the system's predictions significantly rely on gender-specific words and phrases. The authors also argue that eliminating gender-specific information from word embedding can reduce such biases. They consequently discovered pathways for each layer of BERT that predominantly encode gender information. The area created by these directions in the semantic space of word embeddings is known as the gender subspace. The authors also provided one primary direction for each BERT layer to detect fine-grained gender directions. This prevents other vital information from being overlooked and does away with the requirement that gender subspace is realized in several dimensions. According to experiments, such systems operate better when components are not embedded. According to experiments, removing embedding components that point in these directions significantly minimizes the bias caused by BERT in downstream tasks.

\subsection{Contextual Addition}

\begin{figure}[ht]
    \centering
    \includegraphics[width = \linewidth]{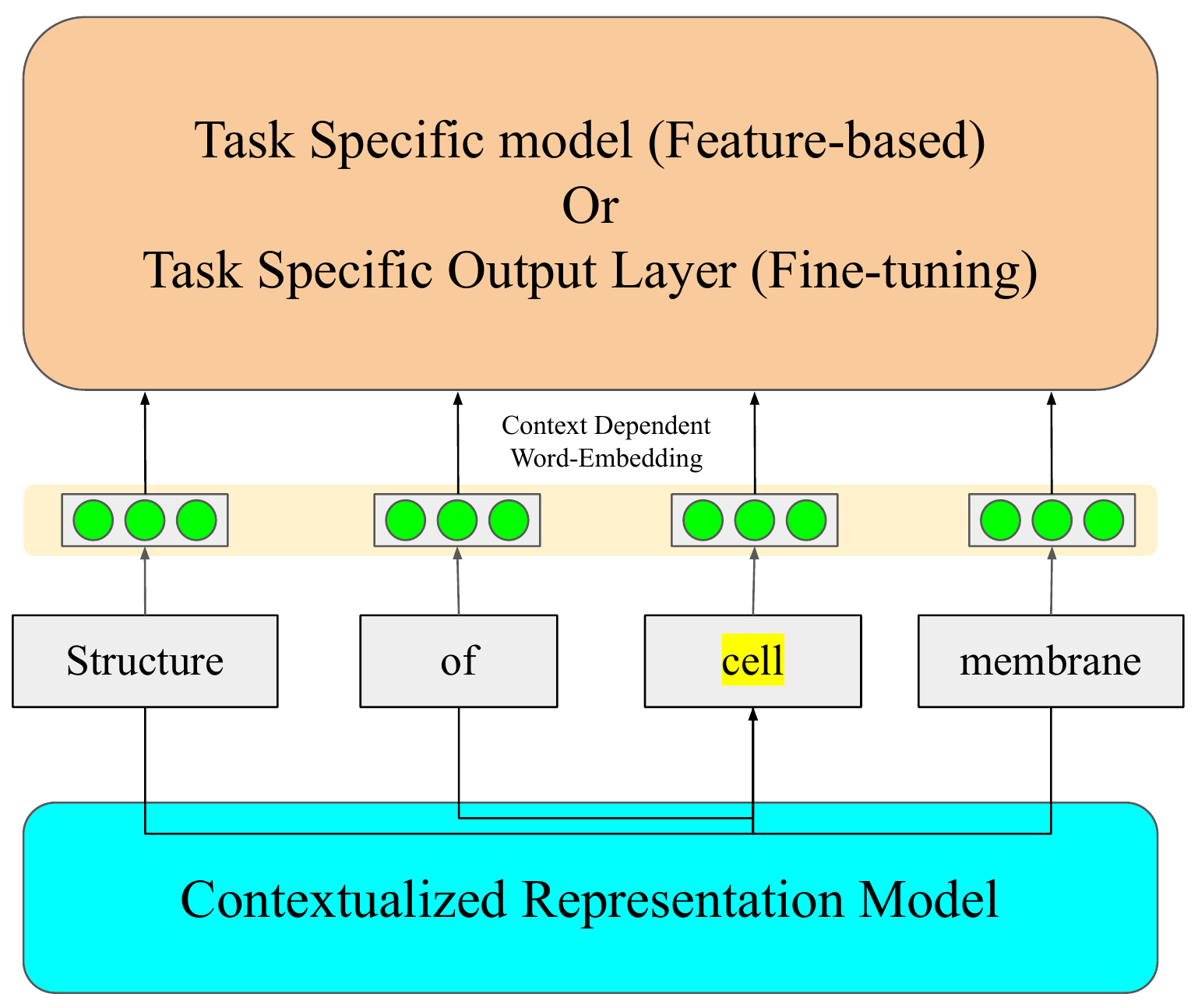}
    \caption{Contextual Addition on Task Specific Model}
    \label{fig:CA}
\end{figure}

\textbf{Basta et al. \cite{basta2020towards}} conducted a study to determine whether recently proposed MT approaches significantly contribute to reducing biases in document-level and gender-balanced data. The authors proposed contextual addition, which is illustrated in Fig. \ref{fig:CA}, and speaker id methodology in a decoder-based NMT model \cite{luo2019encoder}. Their work examined the techniques for incorporating the prior phrase and speaker information into a decoder-based neural MT system, also named WinoMT. Their experiments' architecture only includes the decoder portion of the well-known Transformer, which minimizes training parameters and streamlines the model. The authors noted that WinoMT is a test set lacking speaker identity and information at the document level; therefore, translation using their methodology is carried out without these details. As a result, the system becomes more robust when the information from the previous sentence is included; thus, it is okay for the authors to draw inferences without it. The results show improved translation quality (+1 BLEU point) and gender bias mitigation (+5\% accuracy).

\subsection{Counterfactual Data Substitution}
In addition to traditional word embeddings, evaluating biases contained in their replacements is crucial, according to \textbf{Bartl et al. \cite{bartl2020unmasking}}. By examining connections between gender-denoting target words and profession names in English and German and comparing the findings to actual workforce numbers, they used BERT to quantify gender bias. They also reduced bias by fine-tuning BERT on the GAP corpus after applying Counterfactual Data Substitution (CDS). The gender mitigation technique CDS operates at the corpus level. It operates by randomly applying an intervention to half of a corpus's documents. This intervention aims to maintain the grammatical consistency of a document while inverting all of the gendered terminologies inside it. They demonstrated that while this method of measuring bias works well for languages like English, it does not work well for languages like German, which have a sophisticated morphology and gender-marking system. Their findings highlight the importance of looking at bias and mitigation strategies across languages, especially in light of the current focus on extensive, multilingual language models.

\begin{table*}[htbp]
	\centering
	\caption{Literature review involving the reduction of Gender Bias in Transformer Models}
	\label{tab: methods}
	\resizebox{\linewidth}{!}{
	\begin{tabular}{|c|p{6cm}|p{6cm}|p{6cm}|}
	\hline
	\centering \textbf{Author} &  
    \centering \textbf{Strengths} & 
    \centering \textbf{Weaknesses} &
    \centering \textbf{Notable Findings} \tabularnewline 
    \hline

    \centering Joniak et al. \cite{joniak2022gender} &
    \raggedright 
    1. Identifies weights that, when removed, lead to convergence of debiasing objectives. \\
    2. Combination of debiasing, weight freezing, and movement pruning. \\ 
    3. Enables model optimization by removing bias-related weights. &
    \raggedright 1. Focuses on weight removal without examining other aspects of bias. \\ 2. May affect model performance. &
    \raggedright 1. Gender bias is primarily stored in intermediate layers of BERT \\ 2. A direct relationship between model bias and performance. 
    \tabularnewline 
    \hline

     \centering Bao et al. \cite{bao2019transfer} & 
     \raggedright 
     1. Utilizes knowledge learned from pre-existing models to improve performance on a new task. \\ 
     2. Reduces the need for extensive training data. \\
     3. Enhances performance on tasks requiring less data. &
     \raggedright 
     1. Relies on the availability of pre-existing models. \\ 
     2. May not effectively address task-specific biases. \\
     3. Requires careful fine-tuning to balance general knowledge and task-specific features.&
    \raggedright 
    1. Improvement in coreference system performance on the GAP dataset. \\ 
    2.  Effective reuse of pre-trained BERT knowledge for improved task performance. 
    \tabularnewline 
    \hline

    \centering Vig et al. \cite{vig2020investigating} &
    \raggedright 
     1. Investigates the mechanisms influencing gender bias in models. \\ 
     2. Identifies specialized model components responsible for bias. \\
     3. Examines effects flowing directly and indirectly through mediators. &
     \raggedright 
     1. Limited to specific models (GPT2, XLNet, RoBERTa, DistilBERT). \\ 
     2. Does not provide direct mitigation strategies. \\
     3. Requires careful analysis and interpretation of results.&
    \raggedright 
    1. Gender bias effects concentrated in specific model components. \\ 
    2. Highly specialized behavior observed in certain model components. 
    \tabularnewline 
    \hline

    \centering Bharadwaj et al. \cite{bhardwaj2021investigating} &
    \raggedright 
    1. Evaluates gender bias in CLMs using regression models. \\ 
     2. Identifies gender-specific word dependencies. \\ 
     3. Identifies gender subspace in word embeddings. &
     \raggedright 
     1. Focuses on word embeddings and may not capture all aspects of bias. \\ 
     2.  Does not provide direct mitigation strategies.&
    \raggedright 
    1. CLM predictions are significantly influenced by gender-specific words. \\ 
    2. Identification of gender subspace and specific gender directions in BERT. 
    \tabularnewline 
    \hline

   \centering Basta et al. \cite{basta2020towards} &
   \raggedright 
    1. Incorporates prior phrase and speaker information in NMT models. \\ 
     2. Improves translation quality and reduces gender bias. \\ 
     3. Efficient architecture with reduced training parameters. &
     \raggedright 
     1. Limited to decoder-based NMT models. \\ 
     2. Relies on the availability of previous sentence information. \\ 
     3. May not fully address biases in the source text. &
    \raggedright 
    1. Improved translation quality and reduced gender bias in NMT models. \\ 
    2. Robustness achieved by including previous sentence information.
    \tabularnewline 
    \hline

    \centering Bartl et al. \cite{bartl2020unmasking} &
     \raggedright 
    1. Mitigates bias by fine-tuning BERT on a modified corpus &
     \raggedright 
     1. Effectiveness may vary across languages with complex gender systems. \\ 
     2. It does not work well for languages like German,
which have a sophisticated morphology. &
    \raggedright 
    1. Successful bias reduction in English but challenges in German due to language characteristics.
    \tabularnewline 
    \hline

    \centering Kaneko et al. \cite{kaneko2022gender} &
     \raggedright 
     1. Evaluates bias across languages using English attribute word lists and parallel corpora. &
     \raggedright 
     1. Requires parallel corpora between the target language and English &
    \raggedright 
    1. Identified gender-related stereotypes in MLMs across multiple languages
    \tabularnewline 
    \hline

    \centering De et al. \cite{de2021stereotype} &
    \raggedright 
     1. Introduces measures to analyze and reduce gender bias in contextual language models &
     \raggedright 
     1. Trade-off between skew and stereotype in out-of-the-box models. &
    \raggedright 
    1. Optimized models significantly reduce both skew and stereotype. 
    \tabularnewline 
    \hline

	\end{tabular}}
\end{table*}

\subsection{Multilingual Bilingual Evaluation}

\textbf{Kaneko et al. \cite{kaneko2022gender}} highlighted that existing bias evaluation methods require stereotypical sentence pairings with the same context and attribute terms (e.g., He/She is a nurse). Without requiring manually annotated data, they presented the Multilingual Bilingual Evaluation (MBE) score for assessing bias in several languages using just English attribute word lists and parallel corpora between the target language and English. They evaluated MLMs in eight different languages using the MBE and discovered that all of them include gender-related stereotypes. They manually created datasets for gender bias in Japanese and Russian to evaluate the MBE's validity. The findings show a strong correlation between the gender bias MBE scores and those derived from those mentioned above personally created datasets and the existing English datasets.

\subsection{Skew and Stereotype Methodology}
In a recent study addressing the issue of WinoBias pronoun resolution, \textbf{De et al. \cite{de2021stereotype}} have proposed two measures, namely skew, and stereotype, to analyze and quantify gender bias present in contextual language models. The skew measure aims to reduce stereotypes but may result in increased skew, while the stereotype measure seeks to optimize models using a larger gender-balanced dataset, thereby minimizing both skew and stereotype. The study compared the performance of the optimized BERT model with its unaugmented, fine-tuned counterpart, demonstrating that the optimized model significantly reduces both skew and stereotype. Additionally, the researchers found that out-of-the-box models exhibit a trade-off between skew and stereotype, with RoBERTa and ALBERT-xxlarge displaying reduced skew at the cost of higher stereotypes, while DistilBERT and BERT models have high skew and low stereotypes.

\section{Discussion}
Following the presentation of the formal definitions, detection, and mitigation methods used in this study, we move on to the primary conclusions we draw from it. Four major issues with current studies on gender prejudice come to light, and we examine them below.

\subsection{Gender in NLP}
Researchers often explore the relationship between gender and text using NLP methods. However, it is important to consider the ethical implications of categorizing people based on their gender without thoughtful consideration. Many studies on gender fail to explain how gender labels are assigned, leaving their imputation process open to criticism. Imputing gender labels without proper explanation can be ethically problematic. It is widely recognized that using gender as a binary variable oversimplifies the complexities of gender and can cause harm to non-binary individuals. Researchers should take proactive steps in their methodology to avoid perpetuating damaging prejudices and promote inclusivity. In a recent study by \textbf{Lindqvist et al. \cite{lindqvist2021gender}}, \textbf{it was found that researchers often characterize gender as binary in their research, which can be problematic}. To mitigate these effects, researchers are encouraged to define gender clearly and inclusively in their studies. This means acknowledging and incorporating non-binary gender identities. By extending corpora to include inclusive pronouns and evaluating models' performance on tasks involving non-binary pronouns, researchers can address the limitations of binary assumptions and contribute to a more comprehensive understanding of gender-related phenomena. It is crucial for researchers utilizing transformer models to be aware of the potential biases and limitations in handling gender. By actively incorporating inclusive practices, researchers can avoid marginalizing non-binary individuals and better understand the relationship between gender and text. Ongoing efforts to improve models' performance on non-binary pronouns are essential to ensure fairness and equity in NLP research.

\subsection{Single-Language Focuss}
Gender bias in NLP is a complex issue influenced by cultural and societal perspectives, which can vary across languages and regions. To obtain a more comprehensive understanding of gender prejudice in society, it is crucial to expand research on gender bias beyond the confines of the Anglosphere and include a diverse range of languages and cultures. Unfortunately, previous studies on gender bias have predominantly focused on monolingual research, with a disproportionate emphasis on English or a small subset of high-resource languages. This limitation restricts our understanding of gender bias to a narrow perspective, potentially overlooking the unique manifestations of bias in different linguistic and cultural contexts. To effectively address gender bias in NLP, it is imperative to incorporate a broader range of languages and cultures into research and development efforts. By including languages from diverse regions, we can capture the nuances of gender bias specific to each language and culture. This approach enables a more inclusive and representative analysis of gender bias, fostering a better understanding of its underlying mechanisms and impacts. Furthermore, it is essential to move beyond the default reliance on English-based language models as the primary solution to gender bias. While English may be the most widely studied language in NLP, it should not overshadow the importance of developing language models and solutions tailored to other languages. By prioritizing the development of language models in a variety of languages, we can address gender bias more effectively and provide fair and equitable NLP systems for a broader range of users.

\subsection{Required Formal Testing}
Most articles that have examined gender bias identification in natural language, techniques, or downstream tasks have viewed bias detection as a goal in and of itself or as a way to examine the nature of bias in their interest areas. Commonly used models that have recently produced notable improvements on a variety of NLP tasks did not include any bias research alongside the release. These methodologies typically only undergo post-hoc bias testing after being used in actual applications, which could harm many social groups. Because these models were only tested for gender bias after release, they may have already caused societal harm. We conclude that bias detection should be incorporated into the pipeline for model creation at an early stage and believe that implementing this change will be the main difficulty. Holding academics responsible at the planning stage of research projects—requiring project proposals and publications to include ethical considerations—and subsequently, throughout the peer review process is the best approach to ensure that they adhere to ethical norms. Therefore, extensive and multifaceted bias measures are needed to establish formal testing. We discover that work on gender bias, in particular, suffers from incoherence in evaluating criteria, much like research on social biases. The majority of articles on gender prejudice only take into account one definition of bias and do not conduct enough parallel studies to combine these approaches. One of the key difficulties in this field of research is coming up with a comprehensive definition of gender prejudice because it can be expressed in language in a variety of complex ways. Furthermore, to improve comparability, we strongly advise adopting standardised evaluation benchmarks and examinations.

\subsection{Limited definitions}
In order to establish formal testing, it is necessary to implement extensive and multifaceted measures to address bias. Our research indicates that gender bias studies, much like research on societal biases, suffer from inconsistency in the use of evaluation criteria. Most studies on gender bias focus on only one definition of bias and do not conduct sufficient parallel research to combine various approaches. This presents a significant challenge in defining gender bias comprehensively since it can manifest in language in complex and nuanced ways. To address this, we recommend the development of standardized evaluation benchmarks and tests to improve comparability.

\section{Conclusion and Future Works}
This study sheds light on the issue of gender bias in Transformers by undertaking a comprehensive and critical analysis of various works in this field. Through this analysis, we identify the key challenges and limitations in existing research on gender bias in Transformers and opportunities for further investigation. Our study highlights the need for a more nuanced and sophisticated understanding of gender bias in language models and the development of more effective techniques for detecting and mitigating these biases. By critically evaluating existing research, we are able to identify gaps in our knowledge and suggest potential areas for future inquiry. Overall, our linguistic perspective provides a valuable framework for understanding and addressing the problem of gender bias in Transformers. By focusing on the linguistic processes at play, we can develop more targeted and effective interventions to promote gender-fair language models. There are several potential avenues for future research and practical applications in the realm of gender bias in Transformer models. One area of exploration is the development of more sophisticated and nuanced methods for detecting and mitigating gender bias in language models. This could include the creation of new benchmark datasets, as well as the utilization of advanced machine learning techniques and models. Another potential area of application is the deployment of gender-fair language models in real-world settings, such as chatbots or virtual assistants. Such models could play a critical role in promoting inclusivity and diversity in technology by ensuring that language models are free from gender bias and can interact with users in a fair and equitable manner. Overall, the field of gender bias in Transformer models is rapidly evolving, with new research and applications emerging all the time. By continuing to explore these important issues, we can help build a more just and equitable future for all technology users.

\section{Ethics and Impact Statement}
Our research focuses on the topic of social biases that are inherently present in huge pre-trained transformer models that are widely accessible and employed. Our findings show that bias is a serious issue that the community has to address and that all pre-trained algorithms currently display some sort of biased gender prediction in otherwise neutral circumstances. Our research also depicts the best-proposed solutions to tackle gender bias in transformers.

\bibliographystyle{IEEEtran}
\bibliography{LIPAR}
\end{document}